\documentclass[10pt,twocolumn,letterpaper]{article}

\usepackage[pagenumbers]{iccv} %

\newcommand{\downrightarrow}{\hspace{0pt}%
  \raisebox{1.5pt}{\begin{tikzpicture}[scale=0.4, baseline=(current bounding box.south)]
    \draw[-latex] (0.0, 0.5) -- (0.0, 0.25) -- (0.55, 0.25);
  \end{tikzpicture}} %
}

\usepackage{array}
\usepackage{makecell}
\usepackage{pifont}
\usepackage[section]{placeins}
\usepackage{amssymb}

\newcommand{\mypara}[1]{%
  \textbf{#1}\quad%
}

\usepackage{booktabs} %
\usepackage{tabularx}   %
\usepackage{ragged2e}   %

\usepackage[leftcaption]{sidecap}

\usepackage[accsupp]{axessibility}  %

\usepackage{tikz}

\definecolor{iccvblue}{rgb}{0.21,0.49,0.74}
\usepackage[pagebackref,breaklinks,colorlinks,allcolors=iccvblue]{hyperref}

\def\paperID{15} %
\def\confName{ICCV}
\def\confYear{2025}

\title{Simplifying Traffic Anomaly Detection with Video Foundation Models}

\author{Svetlana Orlova 
\quad Tommie Kerssies
\quad Brun\'o B. Englert
\quad Gijs Dubbelman\\
Eindhoven University of Technology\\
{\tt\small \{s.orlova, t.kerssies, b.b.englert, g.dubbelman\}@tue.nl}
}

\begin{document}
\maketitle
\begin{abstract}
Recent methods for ego-centric Traffic Anomaly Detection (TAD) often rely on complex multi-stage or multi-representation fusion architectures, yet it remains unclear whether such complexity is necessary. Recent findings in visual perception suggest that foundation models, enabled by advanced pre-training, allow simple yet flexible architectures to outperform specialized designs. Therefore, in this work, we investigate an architecturally simple encoder-only approach using plain Video Vision Transformers (Video ViTs) and study how pre-training enables strong TAD performance. We find that: (i) advanced pre-training enables simple encoder-only models to match or even surpass the performance of specialized state-of-the-art TAD methods, while also being significantly more efficient; (ii) although weakly- and fully-supervised pre-training are advantageous on standard benchmarks, we find them less effective for TAD. Instead, self-supervised Masked Video Modeling (MVM) provides the strongest signal; and (iii) Domain-Adaptive Pre-Training (DAPT) on unlabeled driving videos further improves downstream performance, without requiring anomalous examples. Our findings highlight the importance of pre-training and show that effective, efficient, and scalable TAD models can be built with minimal architectural complexity. We release our code, domain-adapted encoders, and fine-tuned models to support future work: \href{https://github.com/tue-mps/simple-tad}{https://github.com/tue-mps/simple-tad}.
\end{abstract}
    
\section{Introduction}
\label{sec:intro}

\begin{figure}[h!]
    \centering
    {
        \hspace*{-2em}
        \includegraphics[scale=0.78]{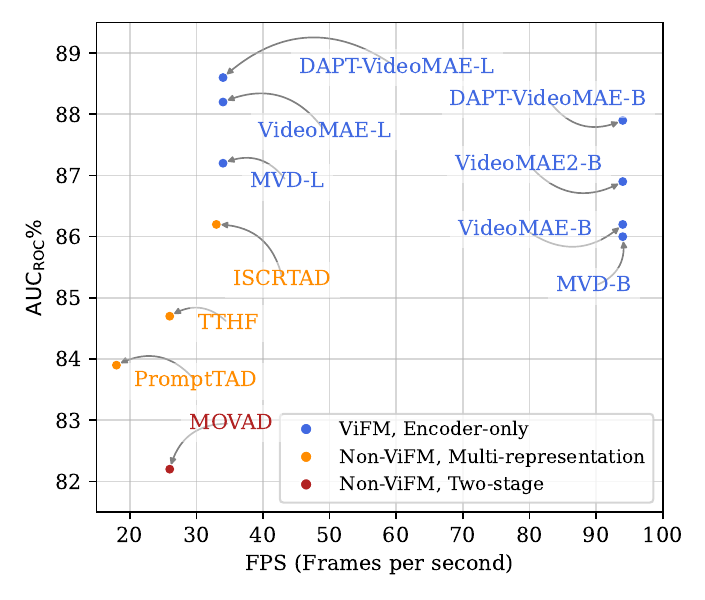}
    }
    \vspace{-2em}
    \caption{\textbf{Traffic Anomaly Detection (TAD) performance on DADA-2000}~\cite{fang2021dada}. Simple encoder-only models with advanced pre-training (blue) are faster and more accurate than recent multi-component architectures (orange and red), see~\cref{tab:performance_comparison_auc,tab:vifms_gen_vs_tad}. %
    }
    \label{fig:eyecatcher}
\end{figure}

\begin{figure}
\centering
\includegraphics[width=\columnwidth]{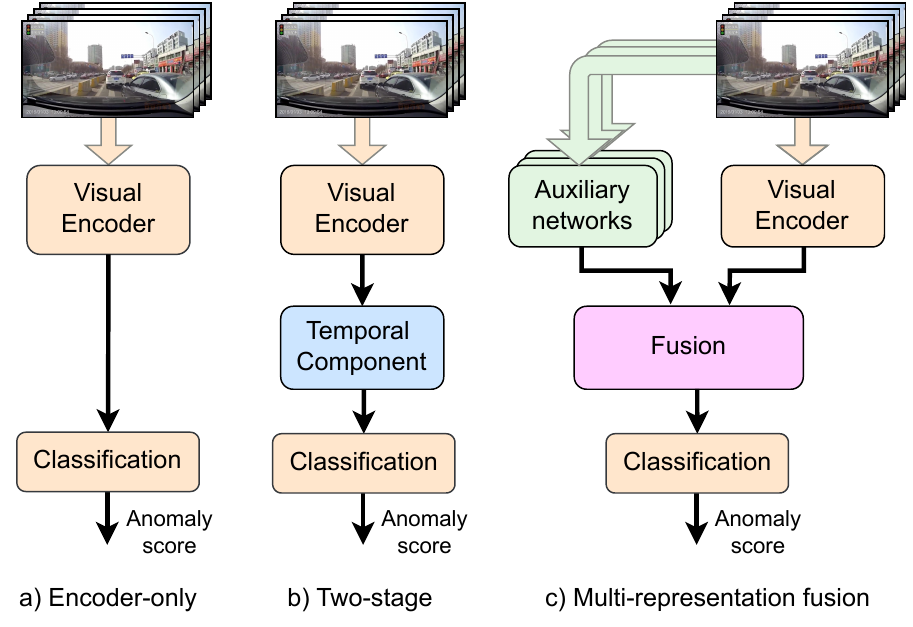}
\caption{Types of model architectures for TAD: simple encoder-only (a), two-stage (b), and multi-representation fusion (c).}
\label{fig:paradigms}
\end{figure}

Traffic risk estimation is fundamental to safe driving, as failures to anticipate danger can lead to life-threatening consequences. Autonomous vehicles must therefore assess potential hazards in real time, even under uncertain, dynamic, and unfamiliar conditions.  
A common formulation for this problem is the ego-centric traffic anomaly detection (TAD) task~\cite{yao2022dota, fang2021dada}, which aims to identify abnormal or dangerous events in a video stream captured by a vehicle‐mounted camera. 
Analyzing the top-performing TAD methods \cite{zhou2022stfe, rossi2024movad, desai2025cyclecrash, liang2024tthf, qiu2025prompttad, liang2025iscrtad}, we find that they rely on specialized, complex architectures illustrated in~\cref{fig:paradigms}, b and c: \textit{two-stage} approaches \cite{zhou2022stfe, rossi2024movad, desai2025cyclecrash}, which combine a vision encoder with a temporal component, and \textit{multi-representation fusion} approaches \cite{liang2024tthf, qiu2025prompttad, liang2025iscrtad}, which fuse additional representations, often generated by separate deep neural networks or model-based algorithms. While these complex designs have improved performance, their impact on efficiency has not been evaluated, even though this is crucial for TAD, where rapid detection is needed to enable timely action and prevent accidents.

Considering that TAD is, in essence, a binary classification task on video, we turn to recent methods for general video classification.
On standard video classification benchmarks \cite{kay2017kinetics, goyal2017something, damen2018epickitchens, soomro2012ucf101}, Video Foundation Models (ViFMs) achieve state-of-the-art performance, predominantly Video Vision Transformers (ViTs)~\cite{dosovitskiy2020vit, arnab2021vivit}, which rely on large-scale self- and weakly-supervised pre-training to learn expressive and transferable spatio-temporal representations, rather than on architectural inductive biases. Recent work in related visual perception tasks shows that advanced pre-training reduces the need for downstream task-specific components~\cite{kerssies2025eomt, vu2024bravo, kerssies2024evaluating}. We hypothesize that the same applies to TAD, such that a simple ViT-based ViFM can be applied to this task and match or even outperform complex architectures. Since TAD depends on motion understanding, we investigate whether pre-training strategies that capture spatio-temporal structure are particularly effective.

To test our hypotheses, we evaluate multiple ViT-based ViFMs, sharing the same plain Video ViT architecture but different pre-training, on two TAD datasets: DoTA~\cite{yao2022dota} and DADA-2000~\cite{fang2021dada}. We adopt an \textit{encoder-only} design, with a single linear layer on top of the Video ViT model, as illustrated in~\cref{fig:paradigms}, a. While prior work has attempted the encoder-only design~\cite{desai2025cyclecrash}, it was found to be inferior~\cite{liang2024tthf, qiu2025prompttad, rossi2024movad}. We revisit this design using stronger pre-training. Similar to prior work, we assess in-domain and out-of-domain generalization, and unlike prior work, we assess the computational efficiency of the models and compare to the best-performing specialized TAD methods. 

We confirm our hypothesis by showing that advanced pre-training enables a plain Video ViT, used as the encoder in an encoder-only design for TAD, to match and even surpass state-of-the-art methods while also being significantly more efficient, as shown in~\cref{fig:eyecatcher} and~\cref{tab:performance_comparison_auc}. Interestingly, performance on standard video classification benchmarks does not correlate with TAD. Our comparison of pre-trained models shows that weak supervision from language and full supervision from class labels are effective on standard benchmarks but less so for TAD, likely because they promote appearance-centric features and semantics that generalize poorly to anomalous motion~\cite{wang2023mvd}. In contrast, self-supervised learning with Masked Video Modeling (MVM), which trains the model to reconstruct missing spatio-temporal regions using both spatial structure and temporal continuity, proves most effective for TAD. Models pre-trained with this objective achieve state-of-the-art performance at their respective model size, as shown in~\cref{tab:vifms_gen_vs_tad}. 

Next, motivated by the scarcity of labeled data for TAD and the abundance of unlabeled driving videos, we explore whether we can leverage self-supervised learning to better adapt an off-the-shelf ViFM to the downstream driving domain.
Specifically, we apply \textit{Domain-Adaptive Pre-Training} (DAPT) \cite{gururangan2020dapt} using the Video Masked Autoencoding (MAE) \cite{tong2022videomae} approach on unlabeled driving videos. We find that MAE-based DAPT, even when applied at a relatively small scale compared to the preceding generic pre-training, significantly boosts the performance even further, particularly for smaller models, as shown in~\cref{fig:exp_dapt_bar_scale}. Importantly, we find that including abnormal driving examples for DAPT is not necessary, as shown in~\cref{tab:dapt_ablation}, which is valuable given the difficulty of collecting them at scale.

We summarize our main contributions as follows:
\begin{enumerate}
    \item We show that a plain encoder-only Video ViT, when equipped with advanced pre-training, outperforms all prior specialized architectures for TAD, while also being significantly more efficient.
    \item We compare pre-training strategies and find that self-supervised learning with a Masked Video Modeling (MVM) objective is most effective, outperforming both weakly- and fully-supervised alternatives.
    \item We demonstrate that Domain-Adaptive Pre-training with MVM leads to measurable performance gains on TAD, even when applied at a small scale to domain-relevant but anomaly-free data.
\end{enumerate}

\vspace{0.5em}
Moreover, since every TAD pipeline relies on its visual encoder, our results offer clear guidance for selecting effective pre-training strategies, enabling future methods to detect traffic anomalies more accurately and robustly.

\section{Related work}
\label{sec:related}

\subsection{Traffic Anomaly Detection}

Traffic Anomaly Detection (TAD) is typically framed as a binary video classification task, where the goal is to detect potentially dangerous or abnormal events in traffic scenarios from the egocentric viewpoint of a vehicle-mounted camera. While related to the broader field of Video Anomaly Detection, which commonly targets static-camera surveillance settings, TAD poses unique challenges due to ego-motion, frequent occlusions, and dynamic interactions of agents in complex driving environments. 
Before the availability of labeled datasets for TAD, earlier approaches often relied on unsupervised reconstruction or prediction of video frames to flag anomalies using temporal autoencoders~\cite{luo2017remembering,wang2018abnormal}, future frame prediction with spatial, temporal, and adversarial losses~\cite{liu2018future}, or spatio-temporal tubelet modeling with ensemble scoring~\cite{yao2022dota}. Some works also explored the use of synthetic data~\cite{schoonbeek2022learning, kim2019crash}.
The introduction of large-scale driving anomaly datasets~\cite{yao2022dota, fang2021dada} with comprehensive annotations has enabled more active development of fully-supervised methods and led to substantial improvements in detection performance. As shown in~\cref{fig:paradigms}, we categorize TAD methods into three classes based on their architectural complexity, which we detail below:

\mypara{Encoder-only design (\cref{fig:paradigms}, a).} Since TAD is a binary classification task, a minimal solution consists of a feature encoder followed by a linear classifier, without any task-specific modules. Prior work~\cite{desai2025cyclecrash} shows that such designs can be effective in related tasks, where R(2+1)D~\cite{tran2018r2p1d} and ViViT~\cite{arnab2021vivit} demonstrate considerably strong performance.
However, recent studies report underwhelming results for encoder-only ablation variants of their methods~\cite{liang2024tthf, qiu2025prompttad, rossi2024movad}.

\mypara{Two-stage design (\cref{fig:paradigms}, b).} Two-stage methods combine a visual encoder with a separate temporal module. VidNeXt~\cite{desai2025cyclecrash} pairs a ConvNeXt~\cite{liu2022convnext} backbone with a non-stationary transformer (NST) to model both stable and dynamic temporal patterns, evaluating and introducing a new dataset CycleCrash~\cite{desai2025cyclecrash} for the related task of collision prediction. Its ablations, ConvNeXt+VT and ResNet+NST, also yield strong results.
MOVAD~\cite{rossi2024movad} uses a VideoSwin Transformer~\cite{liu2022videoswin} as a short-term memory encoder over several frames, followed by an LSTM-based long-term module, achieving state-of-the-art performance for TAD.

\mypara{Multi-representation fusion design (\cref{fig:paradigms}, c).} Fusion-based models, which currently report state-of-the-art performance in TAD, explicitly combine multiple information sources.
TTHF~\cite{liang2024tthf} augments the CLIP~\cite{radford2021clip} framework with a high-frequency motion encoder and a cross-modal fusion module to align motion features with textual prompts. PromptTAD~\cite{qiu2025prompttad} extends MOVAD~\cite{rossi2024movad} by incorporating bounding box prompts via instance- and relation-level attention, enhancing object-centric anomaly localization.
ISCRTAD~\cite{liang2025iscrtad} integrates agent features (e.g., appearance, trajectory, depth) using graph-based modeling and fuses them with scene context through contrastive multimodal alignment for robust anomaly detection.

\vspace{0.5em}
In this work, we follow the simple encoder-only design and explore whether advanced pre-training can compensate for the lack of task-specific architectural inductive biases. 
We demonstrate that, when equipped with rich priors from large-scale self-supervised pre-training, such models can achieve state-of-the-art performance, while remaining architecturally simple and highly efficient.

\subsection{Video Foundation Models}

\begin{table*}
\centering
\footnotesize
\begin{tabular}{llllll} 
\toprule
\textbf{Year} & \textbf{Model} & \textbf{Stage} & \textbf{Type}  & \textbf{Objective} & \textbf{Supervision} \\ 
\midrule
2021 & ViViT \cite{arnab2021vivit} & 1 & FSL & Classification & Class labels \\ 
\midrule
2022 & VideoMAE \cite{tong2022videomae} & 1 & SSL & Masked autoencoding & Video frame pixels \\ 
\midrule
2023 & MVD \cite{wang2023mvd} & 1 & SSL & Masked feature distillation & High-level features of VideoMAE and ImageMAE teachers \\ 
\midrule
2023 & VideoMAE2 \cite{wang2023videomae2} & 1 & SSL & Dual masked autoencoding & Video frame pixels \\ 
 &  & 2 & FSL & Classification & Class labels \\ 
 &  & 3 & FSL & Logit distillation & Logits of larger VideoMAE2 after stage 2 \\ 
\midrule
2023 & UMT \cite{li2023unmasked} & 1 & WSL & Unmasked feature distillation & Features of a vision-language encoder \\ 
\midrule
2024 & InternVideo2 \cite{wang2024internvideo2} & 1 & WSL+SSL & Unmasked feature distillation & Features of VideoMAE2 and a vision-language encoder \\ 
 & & 2 & WSL & Contrastive  & Video-text and audio-text pairs \\ 
 & & 3 & WSL+SSL & Feature distillation & Features of InternVideo2 after stage 2 across multiple depths \\ 
\midrule
2025 & SMILE \cite{thoker2025smile} & 1,2 & WSL & Masked feature distillation & Features of a vision-language encoder \\ 
\bottomrule
\end{tabular}
\caption{\textbf{Overview of ViFMs.} For models trained via distillation, we denote the supervision type(s) used for the teacher. \textbf{FSL}: fully-supervised, \textbf{WSL}: weakly-supervised, \textbf{SSL}: self-supervised learning.}
\label{tab:vitmodels}
\end{table*}

While early 3D CNN-based architectures can be referred to as foundation models~\cite{madan2024foundationvifm}, the term foundation model today typically refers to Transformer-based~\cite{vaswani2017attention} models that leverage large-scale pre-training. For vision, these models typically adopt the Vision Transformer~\cite{dosovitskiy2020vit} (ViT) architecture.
Although other architectures such as recurrent, hybrid, and state-space models are active research areas~\cite{li2024videomamba, yang2022rvit, patraucean2024trecvit}, ViT-based ViFMs are currently the dominant paradigm due to their strong performance, scalability, versatile pre-training strategies, and widespread optimization support (\eg, FlashAttention~\cite{dao2022flashattention}, optimized libraries, hardware acceleration) designed around the plain Transformer architecture.
These qualities, along with the growing availability of pre-trained Video ViTs, make them particularly promising for tasks like TAD, where generalization, robustness, and efficiency are critical.

Unlike convolutional architectures, which impose strong spatial and temporal inductive biases by design, ViTs must learn such priors directly from data during pre-training~\cite{dosovitskiy2020vit}. As a result, their downstream performance depends heavily on the scale and quality of the pre-training dataset, as well as the choice of training objective~\cite{zhai2022scaling}. Pre-training strategies differ in supervision strength and scalability:

\textbf{Fully-supervised learning (FSL)} methods rely on manually annotated labels, providing precise semantic guidance but limited scalability.
First Video ViTs exploited supervised classification as their pre-training method. ViViT~\cite{arnab2021vivit} adopted the ViT architecture to video by introducing spatio-temporal 3D cubes called tubelets instead of 2D patches used for images. While this demonstrated that attention-based models can handle video inputs, the method struggled to balance accuracy and efficiency. 

\textbf{Weakly-supervised learning (WSL)} methods use natural language or metadata as training signals. Though less curated, they offer rich semantic structure and broader concept coverage from web-scale data.
HowTo100M~\cite{miech2019howto100m} introduced large-scale WSL pre-training by aligning video instructions with narrations using contrastive loss, which MIL-NCE~\cite{miech2020end} extended with Multiple Instance Learning (MIL) to handle temporally imprecise supervision. 
UMT~\cite{li2023unmasked} and SMILE~\cite{thoker2025smile} leverage feature distillation from a vision-language model; UMT adopts an unmasked token alignment, while SMILE introduces motion-guided masking for better semantic and temporal understanding.

\textbf{Self-supervised learning (SSL)} methods learn from the data itself without any annotations, enabling large-scale pre-training and highly transferable representations. Unlike FSL and WSL, which introduce annotation noise and bias, SSL provides denser and more unbiased training signals~\cite{chen2020simclr, he2022imagemae}. 
VideoMAE~\cite{tong2022videomae} adopted masked autoencoding (MAE)~\cite{he2022imagemae}, a type of Masked Video Modeling (MVM), as an effective and efficient self-supervised pre-training strategy for plain video ViTs. Its tube masking, when applied to a large fraction of input patches, forces the model to infer spatio-temporal structure from limited visible content. This approach yielded strong results while maintaining architectural simplicity and has since inspired a series of ViT-based Video Foundation Models (ViFMs) that employ self-supervised pre-training~\cite{wang2023mvd, wang2023videomae2, sun2023mme, huang2023mgmae, salehi2024sigma, thoker2025smile}.

\vspace{0.5em}
To our knowledge, we are the first to research which type of video pre-training is most effective for the TAD task, and hypothesize that self-supervised pre-training with an MVM objective, with its emphasis on learning patch-dense and temporally-aware representations, is particularly well-suited for this task. 

\section{Methodology}
\label{sec:methodology}

We fine-tune multiple Video Foundation Models (ViFMs) for TAD. We follow the encoder-only design and attach a single linear layer as a classification head to the output of the encoder. This minimal design ensures that performance primarily reflects the effectiveness of the ViFM backbone in capturing patterns relevant for traffic anomaly detection.

Given a video input \( X_t \in \mathbb{R}^{T \times H \times W \times 3} \) at time \(t\), we extract spatio-temporal feature vector \( F_t \in \mathbb{R}^E \) using the encoder (ViFM), where \(T\), \(H\), and \(W\) are the number of frames, height, and width, and \(E\) is the feature dimension. Then, a linear layer maps \( F_t \) to classification logits \( L_t \in \mathbb{R}^{C} \), with \( C = 2 \). We fine-tune using cross-entropy loss and apply softmax during inference to obtain anomaly class probability as the anomaly score, which can then be binarized with a decision threshold (typically 0.5). More details in~\cref{sup:sec:implementation}.

We investigate (i) whether a simple Video ViT model, pre-trained at scale, can achieve state-of-the-art performance on TAD, (ii) whether better ViFMs for general video recognition are also better for TAD, and what type of pre-training is more effective, and, finally, (iii) whether small-scale domain-adaptive pre-training (DAPT) is feasible and effective for adapting Video ViTs to the driving domain. 

\subsection{Task definition}

We formulate Traffic Anomaly Detection (TAD) as a binary classification task, specifically focusing on frame-level, ego-centric anomaly classification, where each frame captured from a moving vehicle-mounted camera is assigned an anomaly label.

Let $X_t = \{I_{t-\tau+1}, I_{t-\tau+2}, \ldots, I_t\}$ denote a time-ordered sliding window of $\tau$ consecutive video frames captured from a vehicle-mounted camera up to time $t$. Each $I_k$ represents an RGB frame at time step $k$, from the egocentric viewpoint of the vehicle.

The task is to learn a function $f_\theta$ that maps an input window $X_t$ to a prediction $A_t$ at each timestep $t$:
\[
A_t = f_\theta(X_t), \quad t = T, T+1, \ldots
\]
where $A_t \in \{0, 1\}$ is a binary label that indicates whether an anomaly is observed at time $t$.

In the general case, $\tau$ can be 1 and $f_{\theta}$ may use recurrence or autoregressive conditioning on past predictions.

\subsection{Evaluation Procedure}

Prior work typically reports the \textit{Area Under the Receiver Operating Characteristic Curve} (AUC\textsubscript{ROC}) as the primary evaluation metric for TAD~\cite{yao2022dota, zhou2022stfe, liang2024tthf, qiu2025prompttad, liang2025iscrtad}, and we adopt it as well for comparison with previous methods. However, because handling data imbalance is especially important in TAD, we follow related work~\cite{wang2024rs2g, kanna2024mcc, tamagusko2022mcc} and also report the Matthews Correlation Coefficient (MCC). MCC considers all entries of the confusion matrix, including true negatives, and better reflects overall performance under class imbalance~\cite{chicco2023mccreplace}. MCC at a given threshold is defined as: 

\vspace{-1em}
{\small
\begin{equation}
\text{MCC} = \frac{\text{TP} \cdot \text{TN} - \text{FP} \cdot \text{FN}}{\sqrt{(\text{TP} + \text{FP})(\text{TP} + \text{FN})(\text{TN} + \text{FP})(\text{TN} + \text{FN})}},
\label{eq:mcc}
\end{equation}
}where TP, TN, FP, and FN denote true positives, true negatives, false positives, and false negatives, respectively. Note that MCC ranges from $-1$ (inverse prediction) to $1$ (perfect prediction), with $0$ indicating random performance, but we show it in the range -100, 100 to improve readability.

To assess discriminative ability independently of the decision threshold, we compute MCC across thresholds in the range \([0,1]\) and report the area under this curve, referred to as the \textit{Area Under the MCC Curve} (AUC\textsubscript{MCC}). 
We also report MCC at a fixed threshold of 0.5 (MCC@0.5).

Beyond metric design, we adopt a broader evaluation protocol that emphasizes both generalization and efficiency. We assess in-domain and out-of-domain performance, as well as efficiency based on parameter count, peak GPU memory usage, and frames per second (FPS).

\subsection{Pre-trained Encoders}

We select a range of recent Video ViT-based ViFMs that represent various pre-training strategies, and apply them to the TAD task; see \cref{tab:vitmodels} for an overview of their pre-training strategies. When possible, we select variants pre-trained on Kinetics-400~\cite{kay2017kinetics} for consistency. For completeness, we also include R(2+1)D~\cite{tran2018r2p1d}, a fully-convolutional encoder, motivated by recent studies showing its competitive performance in the related task of collision anticipation~\cite{desai2025cyclecrash}.

We include ViViT~\cite{arnab2021vivit} and fully-convolutional R(2+1)D~\cite{tran2018r2p1d} to represent fully-supervised pre-training.
VideoMAE~\cite{tong2022videomae}, MVD~\cite{wang2023mvd}, and VideoMAE2~\cite{wang2023videomae2} are selected to evaluate progressively stronger variants of self-supervised pre-training from videos. 
Among weakly-supervised methods, we assess UMT~\cite{li2023unmasked} and SMILE~\cite{thoker2025smile}, both of which align video tokens with CLIP~\cite{radford2021clip} supervision. 
Finally, InternVideo2~\cite{wang2024internvideo2} combines self-supervised learning from videos and weakly-supervised learning from multiple modalities, and is one of the leading models across numerous video benchmarks. 
Together, these models cover a diverse range of pre-training strategies.

\subsection{Domain-Adaptive Pre-Training}

To better align the Video ViT encoder with the driving domain, we adopt the \textit{Domain-Adaptive Pre-Training} (DAPT) strategy, a simple method originally proposed in the field of natural language processing~\cite{gururangan2020dapt}. DAPT introduces an additional pre-training stage of smaller scale between generic pre-training and downstream fine-tuning, using unlabeled data from the target domain.

We apply VideoMAE-based~\cite{tong2022videomae} DAPT as follows:
\begin{itemize}
    \item \textbf{Step 1: Generic pre-training.} As before, we initialize the encoder with an off-the-shelf VideoMAE model pre-trained on large-scale generic video data, mostly unrelated to the driving domain.
    
    \item \textbf{Step 2: Domain-Adaptive Pre-training (DAPT).} We continue pre-training the same model on a medium-sized dataset of unlabeled driving videos using the exact same VideoMAE reconstruction objective:
    \[
    \mathcal{L}_{\text{DAPT}} = \left\| M \odot \left( x - f_{\theta}(x_{\text{masked}}) \right) \right\|_2^2
    \]
    where \( x \) is the input video, \( x_{\text{masked}} \) is the masked input, \( f_{\theta} \) is the encoder-decoder VideoMAE model, \( M \) is the binary mask, and \(\odot\) is element-wise multiplication.

    \item \textbf{Step 3: Fine-tuning on TAD.} As before, we fine-tune the encoder-only model on TAD datasets using the same configuration with a simple linear classification head.
\end{itemize}

\vspace{0.5em}
The intermediate DAPT step (Step 2) specializes the model towards the driving domain without requiring any labels. It introduces no additional parameters, preserves model efficiency, and remains fully compatible with standard VideoMAE pipelines.

\section{Experiments}
\label{sec:experiments}

\begin{table*}
\centering
\small
\begin{tabular}{l cc cc ccc}
\toprule 
 & \multicolumn{2}{c}{\textbf{DoTA} AUC\textsubscript{ROC}, \%} 
 & \multicolumn{2}{c}{\textbf{D2K} AUC\textsubscript{ROC}, \%} 
 & \multicolumn{3}{c}{} \\
\cmidrule(lr){2-3} \cmidrule(lr){4-5}\\[-1.4em] 
\textbf{Method} &
\small{DoTA$\rightarrow$DoTA} & \small{D2K$\rightarrow$DoTA} & 
\small{D2K$\rightarrow$D2K} & \small{DoTA$\rightarrow$D2K} &
\small{\textbf{\# Param}} & \small{\textbf{Peak GPU}} & \small{\textbf{FPS}} \\
\midrule
\addlinespace[0.5ex]  %
\multicolumn{8}{c}{\textbf{Two-stage TAD methods}} \\
\midrule
VidNeXt \cite{desai2025cyclecrash} & 73.9\phantom{${}^\dag$} & 69.3 & 70.1 & 72.4\phantom{${}^\dag$} & 125 M\phantom{${}^\ddag$} & 0.78 GB\phantom{${}^\ddag$} & 27 \\
ConvNeXt+VT \cite{desai2025cyclecrash} & 73.1\phantom{${}^\dag$} & 61.2 & 66.8 & 67.3\phantom{${}^\dag$} & 125 M\phantom{${}^\ddag$} & 0.77 GB\phantom{${}^\ddag$} & 27 \\
ResNet+NST \cite{desai2025cyclecrash} & 74.0\phantom{${}^\dag$} & 70.1 & 71.2 & 72.3\phantom{${}^\dag$} & 24 M\phantom{${}^\ddag$} & 0.19 GB\phantom{${}^\ddag$} & 124 \\
MOVAD \cite{rossi2024movad} & 82.2\phantom{${}^\dag$} & 77.6 & 77.0 & 75.2\phantom{${}^\dag$} & 153 M\phantom{${}^\ddag$} & 1.10 GB\phantom{${}^\ddag$} & 26 \\
\midrule
\addlinespace[0.5ex]  %
\multicolumn{8}{c}{\textbf{Multi-representation fusion TAD methods}} \\
\midrule
TTHF \cite{liang2024tthf} & 84.7\textsuperscript{\dag} & -- & -- & 71.7\textsuperscript{\dag} & 140 M\phantom{${}^\ddag$} & 0.80 GB\phantom{${}^\ddag$} & 26 \\
PromptTAD \cite{qiu2025prompttad} & 83.9\textsuperscript{\dag} & -- & -- & 74.6\textsuperscript{\dag} & 106 M\phantom{${}^\ddag$} & 1.88 GB\phantom{${}^\ddag$} & 18 \\
ISCRTAD \cite{liang2025iscrtad} & 86.2\textsuperscript{\dag} & -- & -- & 82.7\textsuperscript{\dag} & 359 M\textsuperscript{\ddag} & 1.51 GB\textsuperscript{\ddag} & 33\textsuperscript{\ddag} \\
\midrule
\addlinespace[0.5ex]  %
\multicolumn{8}{c}{\textbf{Encoder-only models}} \\
\midrule
R(2+1)D \cite{tran2018r2p1d} & 81.5\phantom{${}^\dag$} & 76.4 & 78.8 & 78.4\phantom{${}^\dag$} & 27 M\phantom{${}^\ddag$} & 0.27 GB\phantom{${}^\ddag$} & 104 \\
\midrule
DAPT-VideoMAE-S (ours) & \textbf{86.4}\phantom{${}^\dag$} & \textbf{81.7} & \textbf{85.6} & \textbf{84.3}\phantom{${}^\dag$} & 22 M\phantom{${}^\ddag$} & 0.16 GB\phantom{${}^\ddag$} & 95 \\
DAPT-VideoMAE-B (ours) & \textbf{87.9}\phantom{${}^\dag$} & \textbf{83.5} & \textbf{87.6} & \textbf{85.8}\phantom{${}^\dag$} & 86 M\phantom{${}^\ddag$} & 0.54 GB\phantom{${}^\ddag$} & 94 \\
DAPT-VideoMAE-L (ours) & \textbf{88.4}\phantom{${}^\dag$} & \textbf{85.0} & \textbf{88.5} & \textbf{86.6}\phantom{${}^\dag$} & 304 M\phantom{${}^\ddag$} & 1.80 GB\phantom{${}^\ddag$} & 34 \\
\bottomrule
\end{tabular}
\caption{\textbf{Traffic Anomaly Detection (TAD) performance and efficiency.} Video ViT-based encoder-only models set a new state of the art on both datasets, while being significantly more efficient than top-performing specialized methods. FPS measured using NVIDIA A100 MIG, $\frac{1}{2}$ GPU.
\textsuperscript{\dag}From prior work. \textsuperscript{\ddag}Optimistic estimates using publicly available components of the model. ``\textbf{A$\rightarrow$B}'': trained on A, tested on B; \textbf{D2K}: DADA-2000.}
\label{tab:performance_comparison_auc}
\end{table*}

\subsection{Experimental setup}

\mypara{Datasets.}
We evaluate on DoTA~\cite{yao2022dota} and DADA-2000~\cite{fang2021dada}, two large-scale real-world driving anomaly datasets with temporal and frame-level annotations. We also manually refine 1\% of annotations of the DoTA dataset and refer to this variant as DoTA\textsubscript{ref}, more details in~\cref{sup:sec:refining_dota}. DAPT uses Kinetics-700~\cite{carreira2019k700}, BDD100K~\cite{yu2020bdd100k}, and CAP-DATA~\cite{fang2022capdata}.

\mypara{Model input.}
All Video ViTs and R(2+1)D are trained on sliding windows of size 224×224×16 at 10 FPS (1.5s temporal context) by default. For InternVideo2 and UMT, which use tubelets of size 1, we use 224×224×8 at 5 FPS to match the same duration. MOVAD processes videos frame-by-frame at resolution 640×480.

\mypara{Fine-tuning.}
With all Video ViTs, R(2+1)D, and VidNeXt variants, we closely follow the VideoMAE fine-tuning recipe for HMDB51. We train for 50 epochs (5 warmup), with 50K randomly sampled examples per epoch and a batch size of 56. For MOVAD, we follow the original training settings.

\mypara{Domain-adaptive pre-training (DAPT).}
We apply the VideoMAE pre-training strategy~\cite{tong2022videomae}, masking 75\% of tokens, using MSE loss on masked tokens only. Training uses a batch size of 800 and 1M samples per epoch, with 12 epochs. We explore DAPT on three domains: (a) Kinetics-700, (b) BDD100K (normal driving), and (c) BDD100K + CAP-DATA (a mix of normal and abnormal driving). More details can be found in~\cref{sup:sec:implementation}.

\subsection{Can an encoder-only model outperform specialized TAD methods?}

To answer this question, we evaluate models along three critical axes: classification performance, generalization, and efficiency, as shown in~\cref{tab:performance_comparison_auc}. We select a range of recent top-performing methods proposed for the TAD task. Among encoder-only models, we apply the R(2+1)D~\cite{tran2018r2p1d} model and different sizes of VideoMAE pre-trained Video ViTs (with DAPT, see~\cref{sec:dapt}). 

The results show that these Video ViTs consistently strike a strong balance, demonstrating a good combination of predictive accuracy, generalization across datasets, and computational efficiency. Notably, Video ViTs with advanced pre-training achieve the highest AUC\textsubscript{ROC} scores across both DoTA and DADA-2000 datasets, both in-domain and in cross-dataset evaluation, while being highly efficient with a low memory footprint. In contrast, specialized TAD-specific models not only demonstrate lower classification performance but also incur substantially higher computational costs and latency. R(2+1)D and ResNet+NST, while being highly efficient, fall short in predictive quality. This confirms that we can outperform specialized, multi-component TAD methods with a simple encoder-only model by applying a Video ViT with advanced pre-training. 
A more detailed comparison is provided in~\cref{sup:sec:comparison_sota}.

\subsection{What pre-training is better for TAD?}
\label{sec:sub:pretraining}

We evaluate a range of publicly available Video ViT models of several sizes, using their Top-1 accuracy on the general benchmarks Kinetics-400~\cite{kay2017kinetics} and Something-SomethingV2~\cite{goyal2017something} alongside AUC\textsubscript{MCC} and MCC@0.5 on the TAD benchmarks DoTA~\cite{yao2022dota} and DADA-2000~\cite{fang2021dada}. 
Results are shown in \cref{tab:vifms_gen_vs_tad}, from which we observe three key trends.

First, we find that for TAD, self-supervised pre-training, \ie the Masked Video Modeling (MVM) objective dominates: models, pre-trained with the Masked Autoencoding (MAE) approach (VideoMAE), and their distilled variants (VideoMAE2, MVD) achieve the highest AUC\textsubscript{MCC} within each size tier. Additional experiments on different MVM objectives can be found in~\cref{sup:sec:objectives}. 
Second, models that employ weakly-supervised pre-training (UMT, SMILE and InternVideo2) perform worse on TAD, indicating that vision-language supervision may not transfer well to fine-grained temporal anomaly understanding.
Finally, classification accuracy on general benchmarks is not representative of TAD performance: ViViT-Base, despite matching VideoMAE on Kinetics-400, demonstrates significantly lower AUC\textsubscript{MCC}, and InternVideo2, state-of-the-art on general benchmarks, clearly underperforms on TAD. This suggests that the representations that are beneficial for general video classification may not align well with those needed for TAD. In particular, TAD appears to benefit more from dense representations that emphasize fine-grained temporal irregularities rather than the coarse semantic categories typically targeted by general video recognition models. 

The overall top-ranking model on TAD is VideoMAE2, which incorporates dual masking, an additional pre-training step with distillation from a larger model, and $\sim$6 times larger-scale pre-training datasets, compared to other MVM pre-trained models. This confirms that both the scale of pre-training and the choice of objectives significantly impact the transferability of ViFMs to TAD.

\begin{table*}[]
\centering
\small
\begin{tabular}{lllcccccc}
\toprule
 &  &  & \multicolumn{2}{c}{Top-1 accuracy} & \multicolumn{2}{c}{MCC@0.5} & \multicolumn{2}{c}{AUC\textsubscript{MCC}} \\
\cmidrule(lr){4-5} \cmidrule(lr){6-7} \cmidrule(lr){8-9} \\[-1.4em]
\textbf{Model} & \textbf{Variant} & \textbf{Type} & \textbf{K400} & \textbf{SthSthV2} & \textbf{DoTA\textsubscript{ref}} & \textbf{D2K} & \textbf{DoTA\textsubscript{ref}} & \textbf{D2K} \\
\midrule
VideoMAE\textsubscript{1600} \cite{tong2022videomae} & Small & SSL & 79.0 & 66.8 & 55.5 & 49.5 & 52.1 & 48.1 \\ 
MVD\textsubscript{fromB} \cite{wang2023mvd} & Small & SSL & 80.6 & 70.7 & 56.2 & 49.8 & 50.0 & 48.1 \\ 
MVD\textsubscript{fromL} \cite{wang2023mvd} & Small & SSL & 81.0 & 70.9 & 56.5 & 51.1 & 50.2 & 49.1 \\ 
VideoMAE2 \cite{wang2023videomae2} & Small & SSL+FSL & 83.7 & -- & \textbf{56.8} & \textbf{51.6} & \textbf{55.2} & \textbf{50.3} \\ 
InternVideo2 \cite{wang2024internvideo2} & Small & WSL+SSL & \textbf{85.4} & \textbf{71.6} & 51.6 & 44.5 & 49.7 & 43.7 \\ 
\midrule
ViViT \cite{arnab2021vivit} & Base & FSL & 79.9 & -- & 30.7 & 27.6 & 28.9 & 26.7 \\ 
VideoMAE\textsubscript{800} \cite{tong2022videomae} & Base & SSL & 80.0 & -- & 58.0 & 52.0 & 54.5 & 51.2 \\ 
VideoMAE\textsubscript{1600} \cite{tong2022videomae} & Base & SSL & 81.0 & 69.7 & 58.7 & 52.6 & 56.0 & 52.2 \\ 
MVD\textsubscript{fromB} \cite{wang2023mvd} & Base & SSL & 82.7 & 72.5 & 57.8 & 51.6 & 56.0 & 50.9 \\ 
MVD\textsubscript{fromL} \cite{wang2023mvd} & Base & SSL & 83.4 & 73.7 & \textbf{59.2} & 52.1 & \textbf{57.0} & 51.0 \\ 
SMILE \cite{thoker2025smile} & Base & WSL & 83.4 & 72.5 & 56.7 & 48.6 & 54.8 & 49.8 \\ 
VideoMAE2 \cite{wang2023videomae2} & Base & SSL+FSL & 86.6 & \textbf{75.0} & 58.4 & \textbf{54.8} & 56.5 & \textbf{53.4} \\ 
UMT~\cite{li2023unmasked} & Base & WSL & 87.4 & 70.8 & 48.4 & 40.2 & 46.3 & 38.1 \\ 
InternVideo2 \cite{wang2024internvideo2} & Base & WSL+SSL & \textbf{88.4} & 73.5 & 52.2 & 44.2 & 50.0 & 43.1 \\ 
\midrule
VideoMAE\textsubscript{1600} \cite{tong2022videomae} & Large & SSL & 85.2 & 74.3 & \textbf{61.6} & \textbf{56.9} & \textbf{59.7} & \textbf{55.4} \\ 
MVD\textsubscript{fromL} \cite{wang2023mvd} & Large & SSL & \textbf{86.0} & \textbf{76.1} & 60.5 & 54.6 & 59.0 & 53.7 \\ 
\bottomrule
\end{tabular}
\caption{\textbf{Comparing Video ViT pre-training strategies.} In contrast to general video classification benchmarks (K400, SthSthV2), fully and weakly supervised pre-training are less effective for TAD (DoTA\textsubscript{ref}, D2K), while self-supervised pre-training yields the best performance.
\textbf{FSL}: fully-supervised; \textbf{WSL}: weakly-supervised; \textbf{SSL}: self-supervised learning; \textbf{K400}: Kinetics-400~\cite{kay2017kinetics}; \textbf{SthSthV2}: Something-SomethingV2~\cite{goyal2017something}; \textbf{D2K}: DADA-2000~\cite{fang2021dada}.
\vspace{-0.8\baselineskip}}
\label{tab:vifms_gen_vs_tad}
\end{table*}

\begin{figure}
\centering
\includegraphics[width=0.9\linewidth]{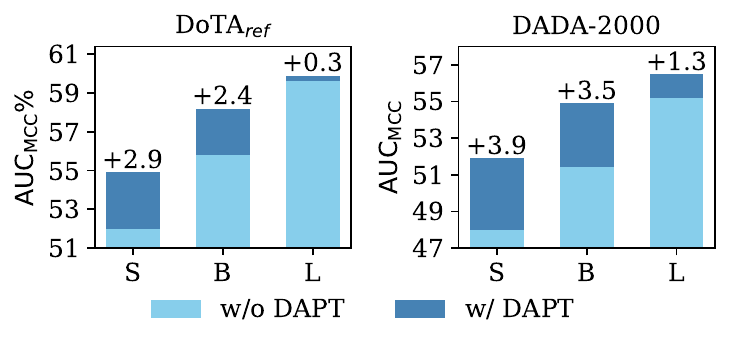}
\caption{\textbf{DAPT scaling across different model sizes.} Smaller models benefit more. \textbf{S}: Small, \textbf{B}: Base, \textbf{L}: Large variants of the Video ViT.
}
\label{fig:exp_dapt_bar_scale}
\end{figure}

\subsection{Domain-Adaptive Pre-Training (DAPT)}
\label{sec:dapt}

Larger ViFMs can be pre-trained on a larger scale and, as a result, exhibit better out-of-the-box generalization across domains, while smaller models have shown to benefit less from longer pre-training due to their limited capacity and faster saturation~\cite{zhai2022scaling}. Therefore, we expect that domain adaptation can help better utilize the capacity of smaller, yet more efficient and faster models. Given that MAE pre-training proves especially effective for TAD, and unlabeled driving data is available in abundance, we investigate whether small-scale self-supervised DAPT with MAE can be an effective and efficient way to scale the performance of smaller models. We initialize a ViT model with VideoMAE pre-trained weights and perform several epochs of additional pre-training with the VideoMAE objective on in-domain data. 
Compared to the original $\sim$192K training steps with batch size 2048, we use only 15K steps with batch size 800.

As shown in~\cref{fig:exp_dapt_bar_scale}, DAPT via MAE brings clear improvements for Small and Base VideoMAE pre-trained models. As expected, given our small-scale DAPT protocol, the Large model sees less improvement.

\begin{table}
\centering
\begin{center}
\begin{tabular}{l@{\hskip 7pt}l@{\hskip 5pt}ll@{\hskip 3pt}l}
\toprule 
 & \multicolumn{2}{c}{\small{\textbf{DoTA\textsubscript{ref}}, AUC\textsubscript{MCC}}} 
 & \multicolumn{2}{c}{\small{\textbf{D2K}, AUC\textsubscript{MCC}}} \\
\cmidrule(lr){2-3} \cmidrule(lr){4-5}\\[-1.4em] 
\textbf{Method} &
\footnotesize{D$\rightarrow$D} & \footnotesize{D2K$\rightarrow$D} & 
\footnotesize{D2K$\rightarrow$D2K} & \footnotesize{D$\rightarrow$D2K} \\
\midrule
w/o DAPT & \small{52.1} & \small{43.8} & \small{48.1} & \small{46.6} \\
Generic DAPT & \small{51.6} \textcolor{red!60!black}{\scriptsize{-0.5}}
            & \small{43.8}
            & \small{48.5} \textcolor{green!60!black}{\scriptsize{+0.4}}
            & \small{46.2} \textcolor{red!60!black}{\scriptsize{-0.4}} \\
Driving DAPT & \small{54.9} \textcolor{green!60!black}{\scriptsize{+2.8}}
                & \small{47.0} \textcolor{green!60!black}{\scriptsize{+3.2}}
                & \small{52.1} \textcolor{green!60!black}{\scriptsize{+4.0}}
                & \small{49.7} \textcolor{green!60!black}{\scriptsize{+3.1}} \\
\downrightarrow + anomalies & \small{55.0} \textcolor{green!60!black}{\scriptsize{+2.9}}
            & \small{47.1} \textcolor{green!60!black}{\scriptsize{+3.3}}
            & \small{52.0} \textcolor{green!60!black}{\scriptsize{+3.9}}
            & \small{49.8} \textcolor{green!60!black}{\scriptsize{+3.2}} \\
\bottomrule
\end{tabular}
\end{center}
\vspace{-0.7em}
\caption{\textbf{DAPT ablation.} Comparing generic (Kinetics-700~\cite{kay2017kinetics}), driving (BDD100K~\cite{yu2020bdd100k}), and driving + anomaly (CAP-DATA~\cite{fang2022capdata}) domains shows that driving videos improve performance without requiring anomalies. Using Video ViT-Small.
``\textbf{A$\rightarrow$B}'': trained on A, tested on B; \textbf{D}: DoTA\textsubscript{ref}; \textbf{D2K}: DADA-2000.}
\label{tab:dapt_ablation}
\end{table}

\begin{figure*}[tbp]
  \centering
  \includegraphics[width=0.95\textwidth]{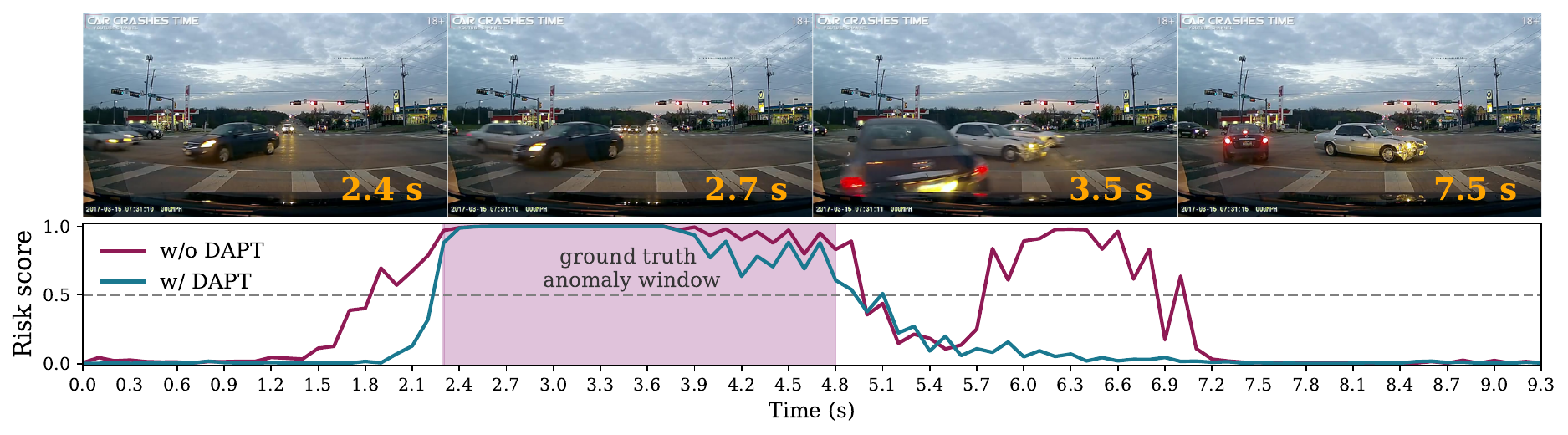}\\[-1ex]
  \includegraphics[width=0.95\textwidth]{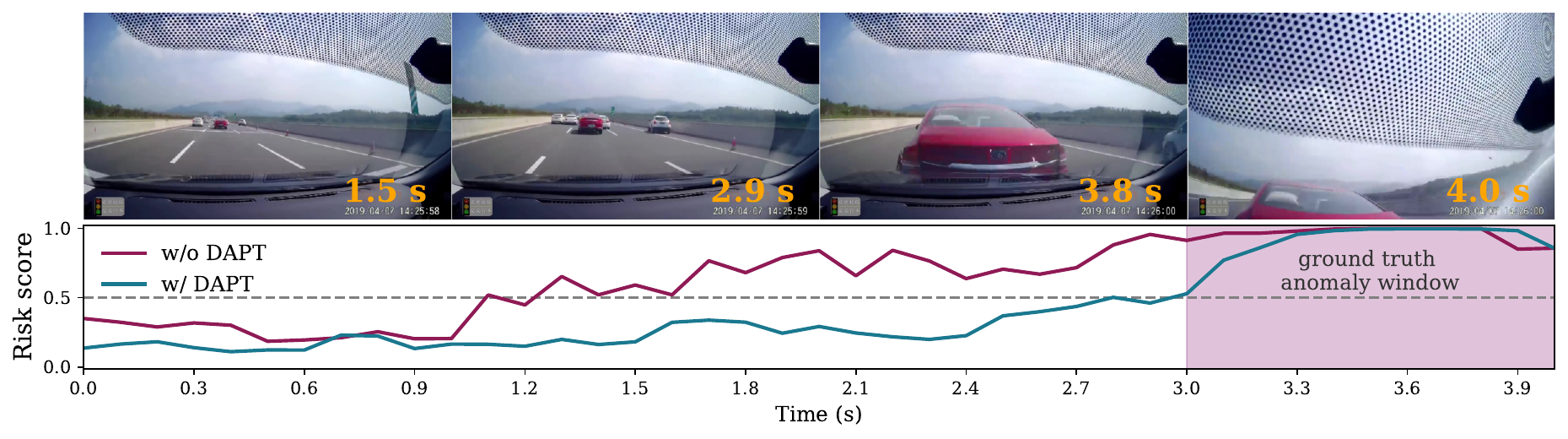}
  \caption{\textbf{Qualitative examples for the effect of DAPT.} 
    Predicted anomaly‐scores of VideoMAE with and without DAPT (top: Video ViT-Small, bottom: Video ViT-Base).}
  \label{fig:dapt_qualitative}
\end{figure*}

To disentangle the impact of domain relevance from that of additional pre-training, we conduct an ablation study on the data used for DAPT. Specifically, we adapt a VideoMAE model, pre-trained on a general human activity dataset, using three types of unlabeled video data: (1) the original, general pre-training domain, which is not related to the TAD task, (2) normal ego-centric driving, and (3) ego-centric driving mixed with anomalies. This setup allows us to evaluate whether the observed improvements stem from domain alignment or from simply continuing pre-training. As shown in \cref{tab:dapt_ablation}, adaptation with domain-relevant data (both normal and abnormal driving) consistently improves in-domain performance and generalization, while additional pre-training on the original, generic domain yields no notable gains. Interestingly, pre-training on normal driving videos is sufficient, and mixing in data with driving anomalies does not provide any further improvement. We conclude that small-scale self-supervised DAPT is a simple and effective way to improve the performance and generalization of smaller Video ViTs for TAD, without the need to collect rare anomalous examples.

Finally, in~\cref{fig:dapt_qualitative} we also include some qualitative examples which clearly demonstrate the positive effect of DAPT.

\section{Discussion}
\label{sec:discussion}

In this work, we show that with advanced pre-training, an encoder-only Video Vision Transformer outperforms all prior Traffic Anomaly Detection models while also being significantly more efficient. However, it remains an open question whether the additional components introduced in earlier work become redundant as pre-training scales, as shown in related perception tasks~\cite{kerssies2025eomt}, or whether they still provide complementary benefits.

\section{Conclusion}
\label{sec:conclusion}

Ego-centric Traffic Anomaly Detection (TAD) is a challenging task that requires modeling motion dynamics and agent interactions. While most recent methods for TAD rely on complex, multi-component architectures, we show that a simple encoder-only design using a plain Video Vision Transformer with strong self-supervised pre-training is not only more efficient, but also more effective and generalizable. Building on this, Domain-Adaptive Pre-Training offers a label-free and compute-efficient way to further boost performance, particularly for smaller models. These findings highlight the strength of learned inductive biases from large-scale pre-training as an alternative to manually crafted architectural complexity, a principle to which TAD is no exception.

Our experiments demonstrate that Masked Video Modeling is the most effective pre-training strategy for TAD, in contrast to general video classification tasks. This suggests that different video tasks may benefit from different pre-training objectives tailored to their downstream requirements. While TAD is a crucial task in autonomous driving (AD), other AD tasks may align more closely with conventional action recognition. This motivates further research into a universally effective video pre-training strategy, evaluated by its generalization across diverse AD tasks. 
We hope our findings provide a foundation for future work in this direction.

\FloatBarrier
\clearpage 
\section*{Acknowledgements}

This work was funded by the Horizon Europe programme of the European Union, under grant agreement 101076754 (project AITHENA). 

We also acknowledge the Dutch national e-infrastructure with the support of the SURF Cooperative, grant agreement no. EINF-10314, financed by the Dutch Research Council (NWO), for the availability of high-performance computing resources and support.

\section*{Disclaimer}

Views and opinions expressed here are those of the authors only and do not necessarily reflect those of the European Union or CINEA. Neither the European Union nor the granting authority can be held responsible for them.

{
    \small
    \bibliographystyle{ieeenat_fullname}
    \bibliography{main}
}

\clearpage
\appendix 

\section*{Appendix}

In this appendix, we provide the following material:
\begin{itemize}
    \item More detailed implementation details (see \cref{sup:sec:implementation}).
    \item Additional comparison between models pre-trained with different Masked Video Modeling (MVM) objectives (see \cref{sup:sec:objectives}).
    \item Detailed evaluation and comparison with existing methods (see \cref{sup:sec:comparison_sota}).
    \item Annotation refinement (see \cref{sup:sec:refining_dota}).
\end{itemize}

\section{Implementation details}
\label{sup:sec:implementation}

\textbf{Encoder-only architecture.} With our simple encoder-only design, we pass a video input \( X_t \in \mathbb{R}^{T \times H \times W \times 3} \) at time \(t\) to a Video Foundation Model (ViFM) encoder, obtaining a spatio-temporal feature vector \( F_t \in \mathbb{R}^E \), and then transform it into classification logits \( L_t \in \mathbb{R}^{C}\) using one linear layer. 

The way \( F_t\) is obtained depends on the specific encoder design. For Video ViTs, we retain the original architectures used for general video classification. First, the input is divided into tubelet tokens (spatio-temporal patches), which are then passed through a series of Transformer blocks. 
Then, the resulting tokens are aggregated into one spatio-temporal embedding. All Video ViT models we used, except for InternVideo2~\cite{wang2024internvideo2}, do not include a dedicated classification token and apply mean pooling over the resulting token embeddings. InternVideo2~\cite{wang2024internvideo2} includes an additional token, but instead of treating it separately, the model aggregates all the resulting tokens using an attention pooling layer, which computes a weighted sum of the embeddings of the token, allowing the model to adaptively focus on the most informative spatio-temporal regions. 
Finally, as in the original designs, the aggregated feature vector is further processed with a final Layer Normalization~\cite{ba2016layer} operation to stabilize and standardize the representation. 
For the R(2+1)D model~\cite{tran2018r2p1d}, which follows a convolutional design, no explicit token aggregation step is required. The model processes the spatio-temporal input through 3D convolutional layers and directly produces a single spatio-temporal embedding. We adopt the implementation from CycleCrash~\cite{desai2025cyclecrash}, where the final spatio-temporal embedding is passed through a batch normalization layer and ReLU activation to obtain the final spatio-temporal feature vector \( F_t\).

\mypara{Domain-adaptive pre-training.} We base our training on the VideoMAE~\cite{tong2022videomae} pretraining recipe with AdamW\cite{loshchilov2017adamw} optimizer and MSE loss on masked tokens. We downscale the training and set the batch size to 800 with 1M samples per epoch, which corresponds to 1250 iterations per epoch. We set the cosine learning rate schedule for 20 epochs with 1 epoch of linear warmup, but stop after 12 epochs by default. We mask out 75\% of the frames.

When mixing normal and abnormal driving data, each batch consists of 480 samples (60\%) from the normal driving dataset BDD100K~\cite{yu2020bdd100k}, and 320 samples (40\%) from the abnormal driving dataset CAP-DATA~\cite{fang2022capdata}. Note that CAP-DATA includes both normal and abnormal driving examples.

\mypara{Fine-tuning.} All Video ViT models are fine-tuned under the same configuration using a batch size of 56.
We closely follow the VideoMAE~\cite{tong2022videomae} fine-tuning recipe for the small HMDB51\cite{kuehne2011hmdb} dataset with AdamW\cite{loshchilov2017adamw} optimizer, cosine learning rate schedule, and the cross-entropy loss. We set the layer decay rate to 0.6, take 50K training examples per epoch (randomly chosen each new epoch), and fine-tune for 50 epochs with 5 epochs of linear warm-up. For the Base and Large variants, we also set the learning rate to 5e-4.

For MOVAD~\cite{rossi2024movad}, we adopt the official publicly available implementation. For VidNeXt~\cite{desai2025cyclecrash}, its ablations, and the R(2+1)D model~\cite{tran2018r2p1d}, we reimplement training within our pipeline to ensure consistency across datasets. Where applicable, we align fine-tuning hyperparameters with the original recipes, while training these models for the same number of iterations as Video ViT models.

\mypara{Evaluation.} We report AUC\textsubscript{ROC} using checkpoints with the highest AUC\textsubscript{ROC} on the validation set. For AUC\textsubscript{MCC} and MCC@0.5, we use checkpoints with the highest AUC\textsubscript{MCC} on the validation set.

\section{Effectiveness of different MVM objectives for TAD}
\label{sup:sec:objectives}

\begin{table*}
\centering
\small
\begin{tabular}{lcccll} 
\toprule
\textbf{Model} & \textbf{AUC\textsubscript{ROC}} & \textbf{AUC\textsubscript{MCC}} & \textbf{MCC@0.5} & \textbf{Masking strategy}  & \textbf{Reconstruction objective} \\ 
\midrule
VideoMAE \cite{tong2022videomae} & 86.3 & 54.8 & 57.8 & Random tube & Pixel \\ 
MME \cite{sun2023mme} & 86.3 & \textbf{55.2} & 57.8 & Random tube & Motion trajectory \\ 
SIGMA \cite{salehi2024sigma} & 86.4 & 54.8 & 57.9 & Random tube & Cluster assignments \\ 
MGMAE \cite{huang2023mgmae} & \textbf{86.6} & 55.0 & \textbf{58.2} & Optical flow guided token & Pixel \\ 
\bottomrule
\end{tabular}
\caption{\textbf{Effect of MVM pre-training objectives.} Despite variations in masking strategy and reconstruction objectives, all MVM-based models show strong performance, indicating that MVM is broadly effective for TAD. Models leveraging motion-aware objectives or masking (MME~\cite{sun2023mme}, MGMAE~\cite{huang2023mgmae}) slightly outperform those relying on raw pixels or semantics (VideoMAE~\cite{tong2022videomae}, SIGMA~\cite{salehi2024sigma}), suggesting that motion modeling benefits TAD, even when pre-trained on general datasets. Using Video ViT-Base models initialized with weights pre-trained on Kinetics-400~\cite{kay2017kinetics} and fine-tuned on DoTA~\cite{yao2022dota}.}
\label{sup:tab:mvm_objectives}
\end{table*}

We demonstrated in~\cref{sec:sub:pretraining} and~\cref{tab:vifms_gen_vs_tad} that among fully-supervised (FSL), weakly-supervised (WSL), and self-supervised (SSL) pre-training, the latter is the most suitable for TAD. Self-supervised pre-training methods for video are predominantly based on the MVM approach, which reconstructs masked video patches. However, specific designs of the MVM pre-training approach vary in how patches are masked and what is reconstructed.

To better understand the effect of specific MVM objectives and masking strategies, we select several ViFMs that differ only in these parameters and fine-tune them on DoTA~\cite{yao2022dota}. All models were pre-trained on Kinetics-400~\cite{kay2017kinetics} for 1600 epochs with the spatio-temporal patch size of 2x16x16 and the same input size of 16x224x224. 

VideoMAE~\cite{tong2022videomae} employs random tube masking, where contiguous spatio-temporal volumes are randomly masked to encourage learning from structured visual dynamics. Being a Masked Autoencoder (MAE) method, it reconstructs RGB pixels of masked patches.
MME~\cite{sun2023mme} introduces motion-aware masked autoencoding by reconstructing dense motion trajectories instead of raw pixels. It randomly masks tubes and trains the model to predict motion features extracted from frame differences, promoting a stronger understanding of dynamic content in videos.
SIGMA~\cite{salehi2024sigma} aims to learn high-level semantics with a projection network, so instead of raw pixels, it reconstructs semantic cluster assignments derived via optimal transport over spatio-temporal features. 
On the opposite, MGMAE~\cite{huang2023mgmae} keeps raw pixels as its reconstruction objective and focuses on advancing the masking strategy. It introduces a motion-guided masking mechanism that leverages optical flow to prioritize masking regions with higher motion, encouraging the model to focus on dynamic and informative parts of the video.

We compare the performance of these models on DoTA in~\cref{sup:tab:mvm_objectives} and see that, despite using different objectives and masking strategies, all MVM-based models achieve similarly strong performance on TAD, confirming that MVM pre-training is robust and broadly effective for this task. Notably, MME slightly outperforms others in AUC\textsubscript{MCC} by predicting motion trajectories rather than pixels, suggesting that incorporating motion dynamics into the reconstruction objective may help the model better capture temporal cues relevant for anomaly detection. MGMAE achieves the highest AUC\textsubscript{ROC} and MCC@0.5, indicating that motion-guided masking can help the model focus on dynamic and informative regions. In contrast, SIGMA, which reconstructs high-level semantic clusters, performs on par with VideoMAE, providing no clear evidence that high-level semantics improve TAD performance. 

These results indicate that motion-oriented objectives and masking strategies provide consistent benefits for TAD, even when pre-training is performed on general video datasets. In contrast, high-level semantic reconstruction shows only marginal gains, suggesting that focusing on such features may not directly benefit tasks requiring fine-grained temporal reasoning like TAD, unless better aligned with the demands of the task.

\section{Detailed evaluation and comparison with existing methods}
\label{sup:sec:comparison_sota}

We present additional experiments that compare our encoder-only ViFM-based models with top-performing specialized TAD methods and analyze key aspects of their performance more closely.

\begin{table}[t]
    \centering
    \small
    \begin{tabular}{@{}lp{7.0cm}@{}}
        \toprule
         \textbf{Short} & \textbf{Anomaly Categories} \\
         \midrule
         ST & Collision with another vehicle which starts, stops, or is stationary\\
         AH & Collision with another vehicle moving ahead or waiting\\
         LA & Collision with another vehicle moving laterally in the same direction\\
         OC & Collision with another oncoming vehicle\\
         TC & Collision with another vehicle which turns into or crosses a road\\
         VP & Collision between vehicle and pedestrian\\
         VO & Collision with an obstacle in the roadway\\
         OO & Out-of-control and leaving the roadway to the left or right \\
         UK & Unknown \\
         \bottomrule
    \end{tabular}
\caption{\textbf{Traffic anomaly categories in the DoTA dataset.} Each category is represented by scenarios with and without ego-car participation.}
\label{sup:tab:dota_categories}
\end{table}

First, we compare our models with existing specialized TAD methods across anomaly categories and groups of the DoTA~\cite{yao2022dota} dataset. DoTA comprises scenarios where the ego-vehicle, from which the video is captured, is either involved in the anomaly or observes other road agents involved in it. We show the list of anomaly categories in DoTA in~\cref{sup:tab:dota_categories}.

We show the results across categories in~\cref{sup:tab:dota_compar_auroc}.
We can see that for most of the categories, and especially for the ego-involved group, our DAPT-VideoMAE models outperform or are on par with specialized methods. 
ISCRTAD~\cite{liang2025iscrtad} and PromptTAD~\cite{qiu2025prompttad} frequently rank among the top three methods across several categories. That suggests that incorporating object information and various related representations, such as depth or high-frequency features, is beneficial, especially in highly untypical accident scenarios (categories UK and UK*).

\begin{table*}
\centering
\small
\begin{tabular}{l *{10}{c}}
\toprule
\multicolumn{11}{c}{\textbf{Ego-vehicle involved video clips}}\\
\midrule
Method 
  & ST & AH & LA   & OC   & TC   & VP   & VO   & OO   & UK & AVG \\
\midrule
STFE~\cite{zhou2022stfe}
  & 75.2 & 84.5 & 72.1 & 77.3 & 72.8 & 71.9 & -- & -- & -- & 75.6 \\
MOVAD~\cite{rossi2024movad}
  & 86.6 & 86.3 & 84.9 & 83.7 & 85.5 & 81.6 & 77.4 & 87.9 & 73.8 & 83.1 \\
TTHF~\cite{liang2024tthf}
  & 86.7 & 90.5 & 89.7 & 87.0 & 89.5 & 77.1 & 87.6 & 90.1 & 70.9 & 85.5 \\
PromptTAD~\cite{qiu2025prompttad} 
  & 84.2 & 90.2 & 88.4 & 85.6 & 89.1 & \textbf{83.6} & 86.8 & 88.7 & 74.6 & -- \\
ISCRTAD~\cite{liang2025iscrtad} 
  & 81.7 & 89.2 & 89.9 & 87.4 & 90.9 & 83.1 & \textbf{90.1} & 88.9 & \textbf{78.9} & 86.7 \\
\midrule
DAPT-VideoMAE-S
  & \textbf{87.3} & \textbf{91.1} & \textbf{90.2} & \textbf{87.6} & \textbf{91.0} & \textbf{83.2} & 85.7 & \textbf{91.2} & 75.8 & \textbf{89.9} \\
DAPT-VideoMAE-B
  & \textbf{87.5} & \textbf{92.0} & \textbf{90.7} & \textbf{89.6} & \textbf{91.8} & 81.6 & \textbf{88.4} & \textbf{91.1} & \textbf{76.1} & \textbf{90.7} \\
DAPT-VideoMAE-L
  & \textbf{87.4} & \textbf{90.6} & \textbf{91.4} & \textbf{88.7} & \textbf{91.8} & \textbf{85.9} & \textbf{89.8} & \textbf{91.6} & \textbf{76.0} & \textbf{90.7} \\
\bottomrule
\toprule
\multicolumn{11}{c}{\textbf{Ego-vehicle NOT involved video clips}}\\
\midrule
 & ST*  & AH*  & LA*  & OC*  & TC*  & VP*  & VO*  & OO*  & UK* & AVG* \\
\midrule
STFE~\cite{zhou2022stfe} 
  & \textbf{80.6} & 65.6 & 69.9 & 76.5 & 74.2 & -- & 75.6 & 70.5 & -- & 73.2 \\
MOVAD~\cite{rossi2024movad}
  & 72.2 & 74.0 & 74.8 & 80.2 & 79.6 & 76.8 & 82.2 & 78.3 & 72.9 & 76.8 \\
TTHF~\cite{liang2024tthf} 
  & 74.9 & 76.0 & 76.4 & 79.8 & 81.5 & 79.2 & 79.0 & 77.5 & 68.9 & 77.0 \\
PromptTAD~\cite{qiu2025prompttad} 
  & 73.8 & \textbf{78.7} & \textbf{81.8} & 82.8 & \textbf{85.1} & 84.6 & 83.1 & 82.4 & \textbf{79.1} & -- \\
ISCRTAD~\cite{liang2025iscrtad} 
  & \textbf{84.6} & \textbf{78.7} & 77.3 & 85.8 & 82.5 & \textbf{86.8} & \textbf{85.4} & \textbf{84.5} & 73.5 & \textbf{82.1} \\
\midrule
DAPT-VideoMAE-S 
  & 78.9 & 77.8 & 78.1 & \textbf{86.6} & 83.9 & 82.9 & 78.6 & 81.8 & \textbf{75.7} & 81.6 \\
DAPT-VideoMAE-B
  & \textbf{80.2} & \textbf{80.0} & \textbf{83.6} & \textbf{86.6} & \textbf{85.8} & \textbf{86.7} & \textbf{84.6} & \textbf{84.4} & 73.0 & \textbf{84.1} \\
DAPT-VideoMAE-L
  & 79.0 & \textbf{81.4} & \textbf{86.1} & \textbf{88.2} & \textbf{86.3} & \textbf{85.8} & \textbf{85.7} & \textbf{85.6} & \textbf{75.6} & \textbf{85.2} \\
\bottomrule
\end{tabular}
\caption{\textbf{Comparison with specialized TAD methods across categories.} Our simple encoder-only models outperform or match top specialized methods across majority of categories. Models using explicit object or scene cues and extra representations (\eg, ISCRTAD~\cite{liang2025iscrtad}, PromptTAD~\cite{qiu2025prompttad}) show advantage in rare or ambiguous scenarios, such as UK and UK*. Reporting AUC\textsubscript{ROC} (\%) of individual accident categories on the DoTA~\cite{yao2022dota} dataset.}
\label{sup:tab:dota_compar_auroc}
\end{table*}
\begin{table*}
\centering
\begin{tabular}{lc@{\hskip 7pt}c@{\hskip 7pt}c}
\toprule
\textbf{Method} & \textbf{Ego\phantom{xxx}} & \textbf{Non‑Ego\phantom{xx}} & \textbf{Both\phantom{xxx}}   \\
\midrule
TTHF~\cite{liang2024tthf}   & 
    80.9\phantom{xxx} & 64.0\phantom{xxx} & 71.7\phantom{xxx} \\
PromptTAD~\cite{qiu2025prompttad}  & 
    79.7\phantom{xxx} & 70.4\phantom{xxx} & 74.6\phantom{xxx} \\
ISCRTAD~\cite{liang2025iscrtad} & 
    82.7\phantom{xxx} & 66.9\phantom{xxx} & 74.2\phantom{xxx} \\
\midrule
VideoMAE-S\textsubscript{HALF} & 
    89.6\phantom{xxx} & 74.8\phantom{xxx} & 81.6\phantom{xxx} \\
VideoMAE-S & 
    90.2\phantom{xxx} & 78.3\phantom{xxx} & 83.7\phantom{xxx} \\
VideoMAE-B & 
    91.2\phantom{xxx} & 79.6\phantom{xxx} & 84.9\phantom{xxx} \\
VideoMAE-L & 
    92.0\phantom{xxx} & 81.0\phantom{xxx} & 86.2\phantom{xxx} \\
\midrule
DAPT-VideoMAE-S\textsubscript{HALF} & 
    90.4\textcolor{green!60!black}{\footnotesize{+0.8}} & 
    77.5\textcolor{green!60!black}{\footnotesize{+2.7}} & 
    83.4\textcolor{green!60!black}{\footnotesize{+1.8}} \\
DAPT-VideoMAE-S & 
    91.2\textcolor{green!60!black}{\footnotesize{+1.0}} & 
    78.7\textcolor{green!60!black}{\footnotesize{+0.4}} & 
    84.3\textcolor{green!60!black}{\footnotesize{+0.6}} \\
DAPT-VideoMAE-B & 
    92.4\textcolor{green!60!black}{\footnotesize{+1.2}} & 
    80.3\textcolor{green!60!black}{\footnotesize{+0.7}} & 
    85.8\textcolor{green!60!black}{\footnotesize{+0.9}} \\
DAPT-VideoMAE-L & 
    91.9\textcolor{red!60!black}{\footnotesize{ -0.1}} & 
    82.3\textcolor{green!60!black}{\footnotesize{+1.3}} & 
    86.6\textcolor{green!60!black}{\footnotesize{+0.4}} \\
\bottomrule
\end{tabular}
\caption{\textbf{Generalization performance.} VideoMAE-based~\cite{tong2022videomae} models generalize significantly better than specialized methods, and even a lightweight variant trained on only half of DoTA~\cite{yao2022dota} outperforms prior work, underscoring the robustness of foundation model pre-training. AUC\textsubscript{ROC} (\%) of different methods trained on DoTA~\cite{yao2022dota} and evaluated on DADA-2000~\cite{fang2021dada}.}
\label{sup:tab:dada_compar_auroc}
\end{table*}

Next, in~\cref{sup:tab:dada_compar_auroc}, we assess the generalization performance of the models by evaluating those trained on DoTA~\cite{yao2022dota} directly on the DADA-2000~\cite{fang2021dada} dataset. VideoMAE-based models already outperform specialized methods by a large margin, and applying DAPT further improves performance. VideoMAE-S\textsubscript{HALF}, a small variant of a Video ViT fine-tuned on only half of the DoTA dataset, already outperforms specialized methods significantly, highlighting the strong generalization capabilities commonly attributed to foundation models.

\section{Annotation refinement}
\label{sup:sec:refining_dota}

While DoTA~\cite{yao2022dota} provides large-scale and comprehensive annotations for traffic anomaly detection, we noticed minor inconsistencies and structural issues that could affect training and evaluation. To identify these cases, we fine-tuned a VideoMAE-Small~\cite{tong2022videomae} model on the official training set and calculated the error rate for each video clip in the set. After that, we flagged the clips with the highest error rate. This model-guided filtering exposed potentially mislabeled or ambiguous samples. Manual review revealed issues, such as substantially imprecise anomaly timing and distinct clips merged into a single video file, which we manually corrected.
We repeated the same procedure (including fine-tuning) on the validation set. Although only about 1\% of the clips were refined, we release the corrected annotations for reproducibility. We use the refined DoTA dataset for comparisons between Video ViT models, and the original dataset when comparing to existing specialized methods.

We repeated the same process for DADA-2000~\cite{fang2021dada}, but a brief manual review did not reveal any significant inconsistencies.

\end{document}